\documentclass[journal,12pt,onecolumn,draftclsnofoot,]{IEEEtran}      
\usepackage{graphicx}
\usepackage{subcaption}
\DeclareCaptionFormat{empty}{}
\usepackage{mwe}
\usepackage{inputenc}
\usepackage{pgfplots}
\pgfplotsset{compat=newest}
\usepackage{amsmath}
\usepackage{amssymb}
\usepackage{algorithm}
\usepackage{nicefrac}
\usepackage[noend]{algpseudocode}
\graphicspath{{}}
\IEEEoverridecommandlockouts                              

\newcommand{\fig}[1]{Fig.~\ref{#1}}

\newcommand{\eqn}[1]{Eq.~\eqref{#1}} 
\newcommand{\alg}[1]{Alg.~\ref{#1}}
\renewcommand{\sec}[1]{Section~\ref{#1}} 

\newcommand{\ie}{i.\,e.~}

\newcommand{\wrt}{w.\,r.\,t.~}
\newcommand{\Real}{\ensuremath{\mathbb R}}        
\newcommand{\T}{\ensuremath{\top}}                
\DeclareMathOperator*{\argmin}{arg\,min}
\DeclareMathOperator*{\argmax}{arg\,max}

\newcommand{\method}{{\textsc{HapDef}}}

\title{\LARGE \bf
Robust Affordable 3D Haptic Sensation via Learning Deformation Patterns}
\author{Huanbo Sun$^{1}$ and Georg Martius$^{1}$
\thanks{*This work was supported by Max Planck Institute for Intelligent Systems}%
\thanks{$^{1}$Huanbo Sun and Georg Martius are with the Max Planck Institute for Intelligent Systems, T\"ubingen, 72076, Germany
        {\tt\small\{huanbo.sun | georg.martius\}@tuebingen.mpg.de}%
}}

\begin{document}
\maketitle
\thispagestyle{empty}
\pagestyle{empty}

\begin{abstract}
Haptic sensation is an important modality for interacting with the real world.
This paper proposes a general framework of inferring haptic forces on the surface of a 3D structure from internal deformations using a small number of physical sensors instead of employing dense sensor arrays.
Using machine learning techniques, we optimize the sensor number and their placement and are able to obtain high-precision force inference for a robotic limb using as few as 9 sensors.
For the optimal and sparse placement of the measurement units (strain gauges), we employ data-driven methods based on data obtained by finite element simulation.
We compare data-driven approaches with model-based methods relying on geometric distance and information criteria such as Entropy and Mutual Information.
We validate our approach on a modified limb of the ``Poppy'' robot and obtain 8 mm localization precision.
\end{abstract}
\section{Introduction}\label{sec:intro}
We are witnessing a rapid development of robot technologies.
Actuators and sensors have become increasingly compact and powerful.
Nevertheless, robots are still far from matching human capabilities especially when it comes to touch sensation.
Haptic information is, however, essential for a reliable interaction with the real world.
It becomes evident that robots need to learn interaction patterns for mastering the real world challenges.
For this, haptic sensors have to be robust in order to sustain long-lasting experiments.
Besides robustness, another important aspect of robotic hardware is its price, availability, and performance.
A low cost makes robotic technologies widely accessible and thus facilitates research.

Currently available \textbf{large area} haptic sensor systems~\cite{Buscher, WANG2015436} are expensive, complicated to integrate, and not robust enough to sustain long-term use.
Array shaped sensors~\cite{tactile3} can localize stimulations but have large amounts of elements and require many wires.
To reduce the hardware complexity, methods such as anisotropic electrical impedance tomography (EIT) ~\cite{Lee} have been proposed.
However, both types are generally not robust because they cover the surface as a skin which makes them vulnerable to impacts with hard or sharp objects.
A tiny crack can destroy the functionality of the entire sensor.
A \textbf{small scaled} sensor BioTac$^{\tiny{\textregistered}}$ \cite{BioTac,BioTac_ML} with integrated multi functionalities has aroused attention since 2007, the functional area is only the finger belly instead of the whole 3D surface due to its structure limitation.
TacTip \cite{TacTip} is an optical tactile sensor with shape size varying from human finger tip to human limb, it is able to detect contacting object shapes accurately while force's information is still not involved in the system.

In this paper, we aim at providing a low-cost, robust and sufficiently precise method for inferring haptic forces on the surface of a 3D structure.
Instead of relying on a dense array of sensors on the surface of the robot, we opt for a small number of physical sensors measuring internal deformations.
This offers a couple of conceptual advantages.
First, the system is robust to environmental impacts because the sensors can be placed inside of the structure. Second, the surface shape can be freely designed. Third, only a few channels have to be read out which reduces both the energy consumption as well as the data rate.

\begin{figure}
      \centering
      \includegraphics[scale=0.8]{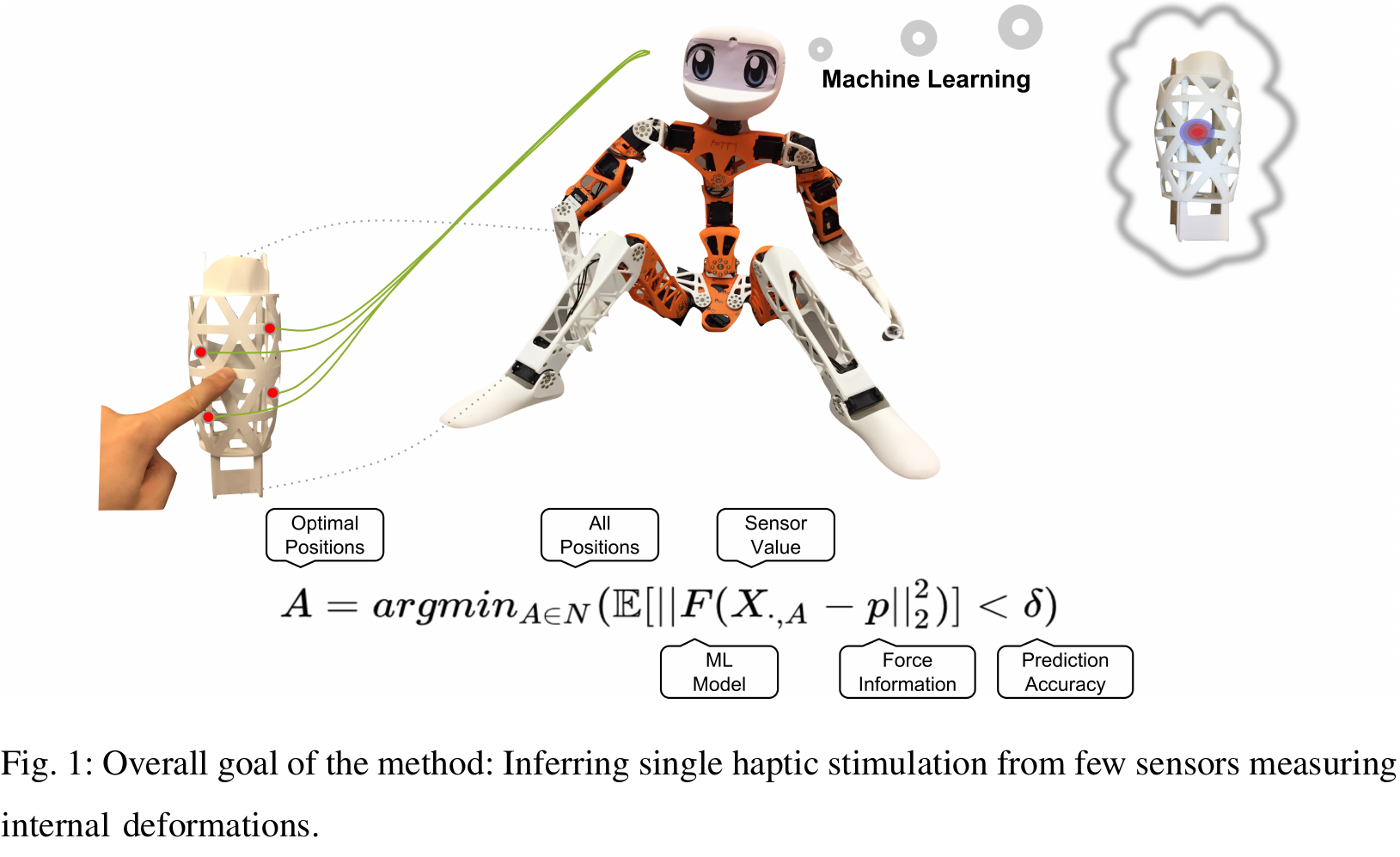}
      \caption{format = empty}
      \label{fig:overview}
\end{figure}
On the downside, a measurement of the sensors does not directly correspond to the impacting force. Instead, an inference mechanism is required to estimate the force.
We propose a data-driven approach using machine learning algorithms to perform this inference efficiently.
In order to require as few sensors as possible, we employ several optimization schemes to determine optimal sensor placement.

The contributions of the paper are as follows.
On the theory side, we:
\begin{itemize}
\item propose a new way of implementing a whole surface haptic sensor,
\item provide a method for determining the optimal number and position of sensors using finite element simulations.
\end{itemize}
On the application side, we
\begin{itemize}
\item provide an assembly method for attaching the strain gauges,
\item designed a hardware system to systematically collect data,
\item implemented the proposed system on a robotic limb.
\end{itemize}

The paper is structured as follows: in \sec{sec:method}, we present the method by first giving an overview and then investigating optimal sensor placement.
In \sec{sec:results}, we present the results on the robotic limb. We close with a discussion in \sec{sec:discussion}.

\section{Method}\label{sec:method}

We propose a method to implement a whole surface haptic force sensor using a minimal amount of deformation sensors (strain gauges) inside of a 3D structure. Using machine learning, the haptic forces are inferred from the few measured deformations, as illustrated in \fig{fig:overview}. We assume that the surface is constructed such that it has a inside rigid support and a flexible outer shell.

In order to place the sensors optimally, we need to get access to data describing how forces applied to the structure propagate into deformations measurable by the sensors.
We do this by finite element simulation.
In our case, the \textsc{Ansys}~\cite{ansys} simulation tool was used. See \fig{fig:datacollection}.

Given the simulated deformation patterns for many force impacts (dataset),  the problem can be stated as follows:
Let $X \in \Real^{N\times M}$ denote the deformation (displacement) of $N$ points on the inside surface of the shell and $p \in \Real^{M \times 3}$ the position of the applied force for $M$ different locations.
We are looking for a subset of location $A$ among all possible points $N$ such that the force locations can we well inferred \ie:
\begin{align}
  \argmin_{A\subset N} \left( \mathbb E[\|F(X_{A,\cdot}) - p\|_2^2] < \delta \right),\label{eqn:goal}
\end{align}
where $F(\cdot)$ is a learned mapping function and $\delta$ is the tolerated error.

Our approach to approximate $\eqn{eqn:goal}$ is composed of the following steps:
\begin{itemize}
\item collect a dataset of deformations from the finite element simulation, see \sec{sec:fes},
\item filter the possible sensor locations according to physical constraints, see \sec{sec:filter},
\item learn a non-linear regression model (SVR) to infer the haptic force positions for unseen stimulation locations, see \sec{sec:inference},
\item select the number and position of sensors needed for a certain pre-determined accuracy, where we evaluate different techniques in \sec{sec:select},
\item validate the prediction quality obtained from differently selected sensor positions, see \sec{sec:placement}.
\end{itemize}

Afterwards, we apply the optimal selected sensor positions and inference model on a real robotic limb (\sec{sec:results}).

\subsection{Finite Element Simulation}\label{sec:fes}
\begin{figure}
      \centering
      \includegraphics[scale=1]{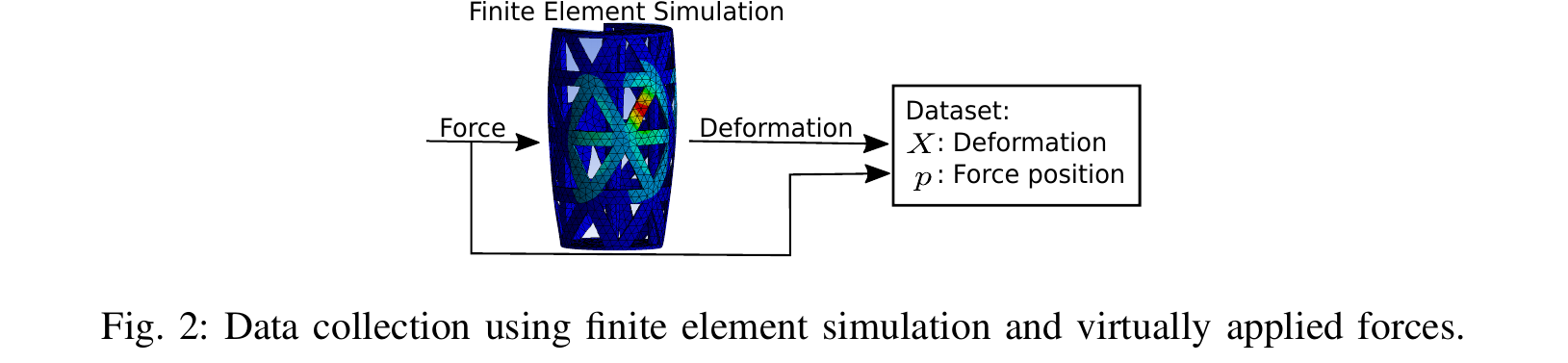}
      \caption{format = empty}
      \label{fig:datacollection}
\end{figure}

In order to get the deformation patterns for a certain impact force, we use a finite element simulation of the 3D structure. The \textsc{Ansys} simulation tool~\cite{ansys} allows importing the 3D-CAD description of the structure to equip with force sensation.
We can apply forces to every location on the surface and record the deformations of all positions on the structure, discretized in a fine manner. The deformation pattern is illustrated in \fig{fig:datacollection}. For this example structure, we obtain around 4000 different force positions and 3000 different sensor positions.

\subsection{Filtering feasible Sensor Positions}\label{sec:filter}
From all positions on the inside of the 3D structure, we need to filter those that allow for physical sensor
 placement.
There are several constraints imposed by the sensor size, their placement restrictions and their
range of detection.
The strain gauge sensors cannot be placed on edges since they need a relatively flat surface.
Placing them near highly rigid support structures is also disadvantageous because only small deformations occur.

In \fig{fig:filtering}(a) the unfolded surface of our example structure is displayed.
The rigid support is at the top and the bottom of the structure, so we discard positions close to that.
In order to get rid of candidate positions at the edges, we use a $k$ Nearest Neighbor (kNN) criterion.
For each candidate position, we consider the center of mass of the neighborhood, see \fig{fig:filtering}(b).
If the center of mass is inside a certain radius, the position is kept (red points in \fig{fig:filtering}(c)).
In our examples, we get around 2100 remaining points.

\subsubsection{Reducing Number of Candidates using Compressive Sensing}\label{sec:compress}

To make the optimal selection of sensor positions more efficient, we further reduce the number of candidates.
Compressive Sensing techniques~\cite{candes2008} can be employed here, which are optimized to reconstruct sparse or compressible signals accurately from a limited number of measurements.
A lossless reduction is not possible in our case, but we can bound the maximal tolerated reconstruction error.

We use PCA with QR-Pivoting~\cite{Manohar2017}.
At this point, we only give an intuitive understanding and elaborate the details below in \sec{sec:pca-qr}.
The method uses PCA to compute the principal components explaining the variance in decreasing order.
The QR-pivoting selects those positions (sensors) that are most important for the top principal components.

\begin{figure}
      \centering
      \includegraphics[scale=1]{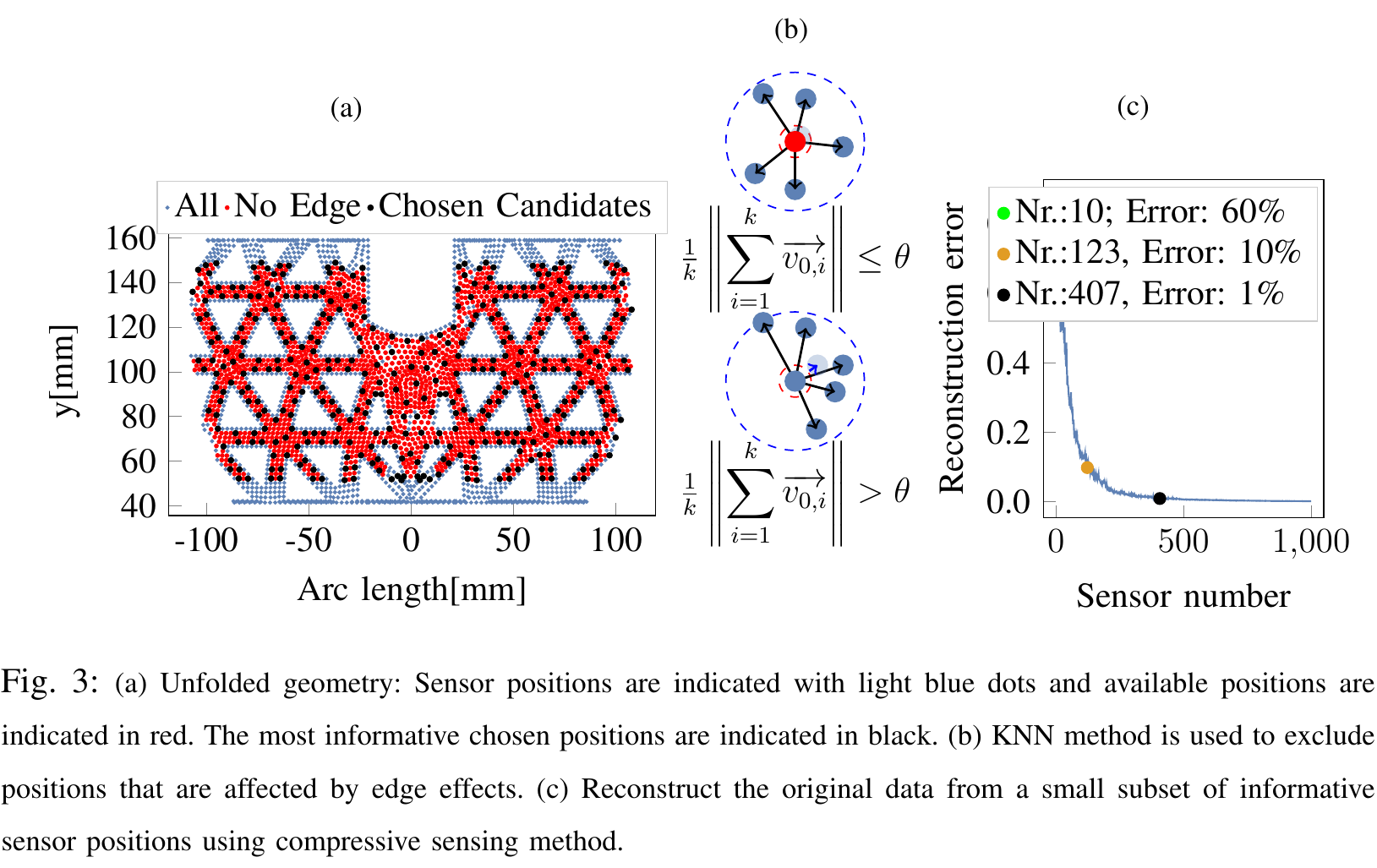}
      \caption{format = empty}
      \label{fig:filtering}
\end{figure}

In \fig{fig:filtering}(c) the linear reconstruction error (unexplained variance) in dependence on
the number of selected sensor positions is displayed.
In our example, for $1\%$ error, we can select 407 out of 2162.
The points are also marked in black in \fig{fig:filtering}(a).

So far, we have automatically selected a set of candidate positions.
\subsection{Force position inference (SVR)}\label{sec:inference}
From the set of candidate positions, the task is now to find a smaller subset that is sufficient to make the inference about deformations anywhere on the structure.
We restrict ourselves in this paper to a method that infers the position of a single impacting force.

Support Vector Machines (SVM) are popular kernel-based algorithms~\cite{Cortes1995} combining the strength of nonparametric techniques with efficient storage requirements of parametrized models.
In this paper, we face a regression task. The Support Vector Regression (SVR) method~\cite{Smola2004} is proper to be applied here.
The idea is that the input $x$ is nonlinearly mapped  into a high-dimensional feature space where a linear regression with maximum margin is performed.

The model is
\begin{align}
F(x, w) &=  \sum_{ i =1 }^{ m }{w_i \cdot g_i(x) +w_0 }\label{eqn:svr:model}
\end{align}
where $g_i(x)$ is the nonlinear feature map, $w$ are the parameters.
The regression error for each example $x_i$ is defined as $\xi_i = \max(0,|F(x_i,w) - p_i| - \epsilon)$,
\ie the deviation from the target $p_i$ larger than $\epsilon$.
In our case, the input $x$ is the vector of deformations at the selected sensor positions and the target $p$ is
 the $p_x,p_y,p_z$ position of the applied force point. In fact, we use one SVR model for each of the target dimensions.

SVR is formulated as a minimization of the following function:
\begin{align}
 L = {1 \over 2 }\| w \|_2 ^2 + C \sum_{i=1}^{n} \xi_i \label{eqn:svr:L}
\end{align}
where the hyper-parameter $C$ controls the trade-off between complexity of the regression model (norm of $w$) and the error.

Interestingly, only scalar products of elements in the feature space are computed, such that one can
directly express the scalar product using an appropriate kernel function $k(x_i,x_j) = g(x_i)^\T g(x_j)$. A popular kernel function is the radial basis function (RBF) or Gaussian kernel:
\begin{align}
k(x_i,x_j)&= \exp\left(- \gamma \left\|x_i -x_j\right\|^{2}\right)\label{eqn:rbf}
\end{align}
with hyperparameter $\gamma$ controlling the sensitivity to distance.
We use the Python \emph{sklearn}~\cite{sklearn} implementation.
We choose RBF kernels and used k-fold cross-validation to select the optimal $C$, $\varepsilon$ and $\gamma$.

As a remark, using all candidate positions and $85\%$ of training data SVR can achieve an average test error of $0.6\,\mathrm{mm}$ ($C = 0.1$, $\varepsilon = 10^{-4}$, $\gamma = 2 \times 10^{-3}$, 5-fold cross-validation).
\subsection{Optimal Sensor Placement}\label{sec:select}

In this section, we propose different ways to select sensor positions, which can be generally grouped into data-driven methods and geometry/model-based methods.

Data-driven methods make use of previously collected data and select the subset of sensors with which
the regression model performs the best.
Geometry/model-based methods do not need access to the measured data.
Instead, they rely only on the geometric position of the sensors.
If sufficient data is available then the data-driven methods can be more accurate because they
have access to the actual dependency between the sensor locations.

In this paper, we propose to use a greedy SVR approach.
As a comparison, we provide also results using a linear compressive sensing method and two geometry-based methods.

\subsubsection{Nonlinear method: Greedy Support Vector Regression}\label{sec:greedysvr}

In principle, we want to select this combination of $K$ sensors that perform best on average at inferring the force position for unseen stimulations.
The problem is that we would need to search through $n\choose K$ different possibilities, which is intractable for $K>4$.
Thus we employ a greedy strategy:
Start with the best single sensor position and then add the second sensor position that gives the best performance and so forth.
Performance is defined in terms of $k$-fold cross-validation, see \alg{alg:greedysvr}.

\begin{algorithm}
\caption{Greedy SVR}\label{alg:greedysvr}
\begin{algorithmic}[1]
  \State {Input data: Deformation: $X \in \Real^{N \times M}$, force position : $y\in \Real^{M \times 3}$, maximum sensor budget $K$}
  \State {Data standardization \& k-fold cross-validation dataset preparation $\{X^1,...,X^k\}$,$\{y^1,...,y^k\}$ }
  \State {Selected nodes: $A=\emptyset$}
  \For {$i:1$ \textbf{to} $K$}
    \For{$m:1$ \textbf{to} $M$ $\not\in A$}

      \For{$j:1$ \textbf{to} $k$}
        \State $A' \leftarrow A \cup m$
        \State $e_j = \|\text{SVR}(X^j\{A'\}) - y^j\|$
      \EndFor
      $error_m = \text{mean}(e)$
    \EndFor
    \State $A \leftarrow A \cup \arg \min(error)$
  \EndFor
\end{algorithmic}
\end{algorithm}

\subsubsection{Linear method: PCA with QR Pivoting}\label{sec:pca-qr}
As mentioned in \sec{sec:compress}, the number of sensors can be reduced using methods from compressive sensing.
In this section, we provide details about the specific method, namely PCA with QR-Pivoting.

Compressive sensing depends on two major functional matrices: feature transform basis $\Psi \in \Real^{N \times N}$ and sub-sampling matrix $\Phi \in \Real^{m \times N}$.
$\Psi$ is designed to transform raw measurements $x \in \Real^{N \times 1}$ into a sparse representative space $\alpha \in \Real^{N \times 1}$ where $\alpha$ has only $k$ nonzero elements:
\begin{align}
x = \Psi  \cdot \alpha
\label{eqn:feature compress}
\end{align}
$\Phi$ subsamples $m$ measurements from $N$ optimally such that the representation $\alpha$ may be most accurately reconstructed from the measurements $\hat{x} \in \Real^{m \times 1}$ using $l_1$ norm:
\begin{align}
\alpha = \arg\min_{\alpha'}  \|\alpha'\|_{1} \hspace{1cm}
\mathrm{s.t.} \hspace{0.5cm} \hat{x} = \Phi \cdot \Psi \cdot \alpha
\label{eqn:compress recon}
\end{align}
\eqn{eqn:compress recon} is a linear optimization problem and conditioned on the operator ($\Phi \cdot \Psi$). The central challenge is to design a optimal $\Psi$ compressing raw data $x$ efficiently and find a good $\Phi$ such that the operator ($\Phi \cdot \Psi$) is well-conditioned.

PCA is a linear unsupervised dimension reduction method \cite{Jolliffe:1986}.
It finds the directions of maximum variance in high-dimensional data and projects data onto a smaller dimensional subspace while remaining most of information.
We keep the first $k$ principle components of $\Psi$ to ensure the $k$ sparsity in $\alpha$ and then select an optimal sub-sampling $\Phi$ to constrain the reconstruction error of \eqn{eqn:compress recon} based on the condition number criterion.
The condition number of the operator ($\Phi \cdot \Psi$) is denoted as:
\begin{align}
c = {{\sigma_{max}(\Phi \cdot \Psi)} \over {\sigma_{min}(\Phi \cdot \Psi)} }
\label{eqn:condition nr}
\end{align}
Since $\Phi$ is a permutation measurement matrix, it can be designed as the column pivoting matrix of $\Psi$.
QR factorization with column pivoting is used to select the highest singular values such that $\sigma_{min}$ is maximized while $c$ is minimized.
Details are shown in \alg{alg:pca-qr}.

\begin{algorithm}
\caption{PCA with QR Pivoting}\label{alg:pca-qr}
\begin{algorithmic}[1]
\State {Input data: Deformation Pattern: $X^{N \times M}$, Maximum Sensor Budget $K$}{}
\State {Data standardization}
\For {$i:1$ \textbf{to} $K$}:
	\State Pick $1:i^{th}$ principle components $\Psi^{N \times i}$
	\State $P  \leftarrow \textbf{QR-Pivoting} (\Psi^{N \times i})$
	\State Sensor node marker = $P[0:i]$
\EndFor
\end{algorithmic}
\end{algorithm}

\subsubsection{Model-based methods using Gaussian Process}
We want to compare the data-driven methods with those relying only on the geometric location of the sensors.
They assume that the geometry is homogeneous and all sensors have a fixed sensing radius, hence they are called model-based.
A convenient model to predict unmeasured sensor values is to use a Gaussian Process (GP)~\cite{cressie1993, MacKay2002}.
It models a distribution over functions with a continuous domain, here functions from sensor location to deformations.
It uses a similarity between locations (sensors) which is measured by a localized kernel to construct a covariance structure.
This in turn is used to make sure that predicted values of similar locations are similar.
However, for a new location, GPs predict not only a mean estimate but also the uncertainty represented by a one-dimensional Gaussian distribution.

\paragraph{Gaussian Process and model/kernel selection}
We start with formalizing the prediction procedure of a GP and then
 choose the right kernel.

Given a set of sensors $A$, their positions $p_A$, and their deformation data $X_{A,\cdot}$
we can predict the distribution of deformations at a different sensor $y$ with location $p_y$.
The mean $\mu_{y|A}$ and standard deviation $\sigma_{y|A}$ are given by
\begin{align}
\mu_{y | A} &=  \mu_y + \Sigma_{y|A} \cdot \Sigma_{AA}^{-1} \cdot (X_{A,\cdot} - \mu_A) \label{eqn:gp:mean}\\
\sigma_{y|A}^2 &= k(y,y)  -  \Sigma_{y|A} \cdot \Sigma_{AA}^{-1} \cdot \Sigma_{y|A}^T\,, \label{eqn:gp:std}
\end{align}
where $\mu_A = \frac 1 {|A|} \sum_{a\in A} X_{\cdot,A}$ is the vector of mean sensor values for the sensors in $A$,
$\Sigma_{AA}(i,j) = k(p_i,p_j) + \beta^{-1}\delta_{i,j}$ is covariance matrix/kernel matrix for all sensors $i,j\in A$ with $\delta_{ij}$ being the Kronecker delta and $\beta$ is a hyperparameter.
Similarly, $\Sigma_{y|A}(i) = k(p_i,p_y)$ is the vector of similarities with the new sensor location.

The GP is a non-parametric process after picking the kernel $k(\cdot,\cdot)$.
Typical choices are polynomial, rational quadratic, and exponential kernels.
In our application, the deformations vary smoothly and locally \wrt the location,
 as they can be described by the bending of a thin plate~\cite{ventsel2001thin}
Thus, we use an exponential kernel
\begin{align}
k(x, x') = \exp\left(- \left(\frac{d(x,x')}{l_{scale}}\right)^{l_p}\right)
\end{align}
with hyperparameters for the length scale $l_{scale}$ and for the distance norm $l_p$.
The distance is measured by $d(x,x')$ which is the approximate geodesic distance instead of the Euclidean distance because the surface of the 3D structure is curved.

\begin{figure}
      \centering
      \includegraphics[scale=1]{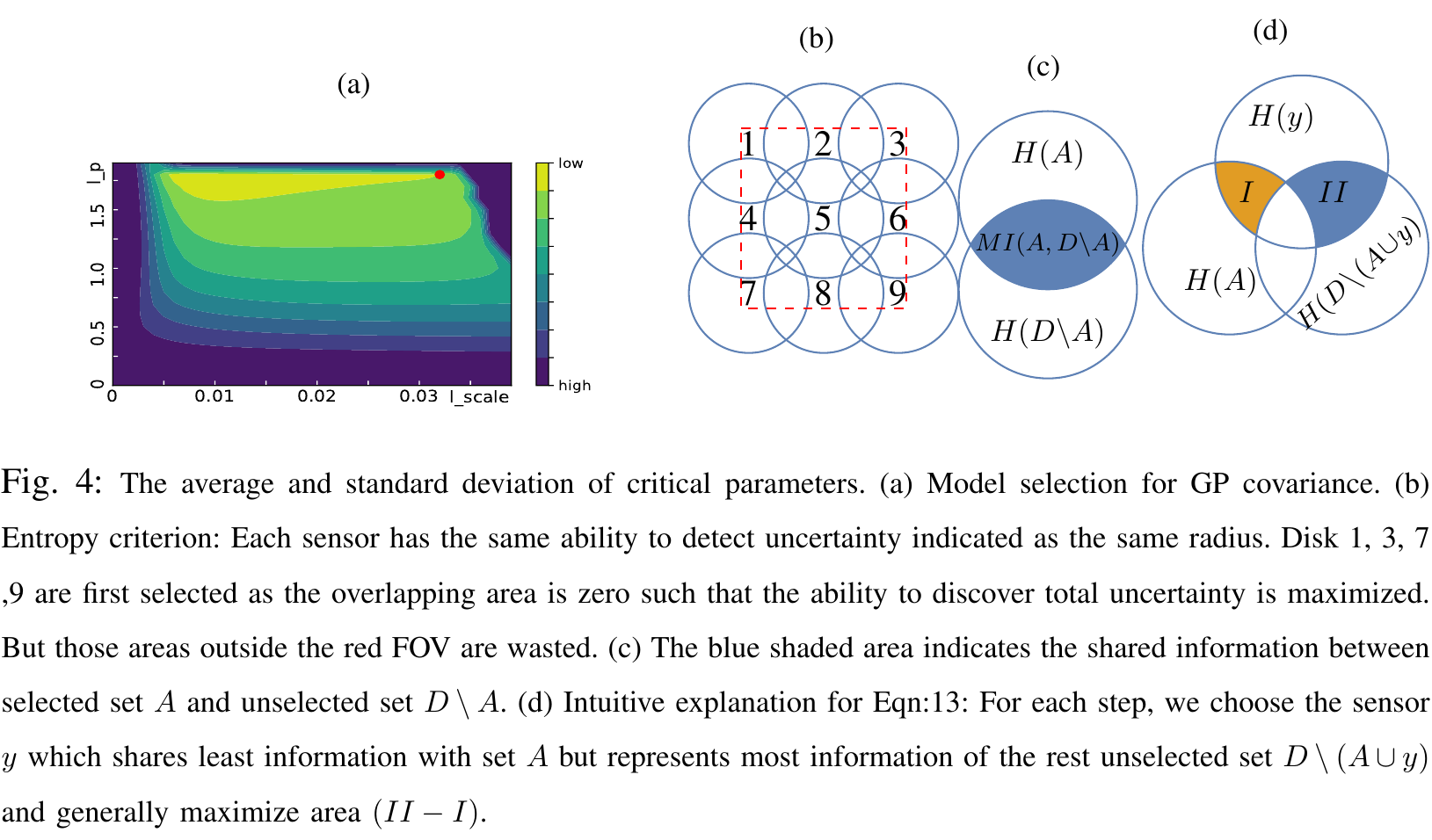}
      \caption{format = empty}
      \label{fig:gp}
\end{figure}

For hyperparameter selection, we use cross-validation~\cite{Rasmussen06} which was shown to be robust.
The grid search for the parameters $l_{scale}$ and $l_p$ is shown in \fig{fig:gp}(a).
We choose $l_{scale} = 0.033$ and $l_p=1.9$.

The probabilistic modeling of the data  allows us to use information criteria for selecting the most informative sensor positions.
The first method minimizes the uncertainty about the non-measured locations (Entropy)
 and the second maximizes the Mutual Information (MI) between selected sensor and the non-measured locations.

As before we select $k$ best sensor positions $A$ out of all permissible locations $V$ (red and black points in \fig{fig:filtering}(a)).

\subsubsection{Pick locations with maximal uncertainty -- Entropy}\label{sec:entropy}
By intuition, a good design is to pick sensor locations that minimize the uncertainty about the entire permissble locations $V$.
This can be quantified by the conditional entropy of the unobserved locations  $V_{\setminus A}$ given the observed ones, \ie $H(V_{\setminus A} | A)$~\cite{krause08near}.
Mathematically, we aim at:

\begin{align}
A^* = \argmin_{A \subset V} H(V_{\setminus A} | A) = \argmax_{A \subset V} H(A)\,,\label{eqn:entropy}
\end{align}
see~\cite{krause08near} for details.
Since \eqn{eqn:entropy} involves a combinatorial search we solve it in a greedy fashion, as done in the SVR case in \sec{sec:greedysvr}.
The entropy of a gaussian distribution is analytically given as $H(\mathbb N(\mu,\sigma^2)) = \frac 1 2\ln 2\pi e \sigma^2$.
The algorithm is detailed in \alg{alg:entropy}

\begin{algorithm}
\caption{Maximum Uncertainty: Entropy}\label{alg:entropy}
\begin{algorithmic}[1]
\State {Input data: GP, sensor budget $K$}{}
\For {$i:1$ \textbf{to} $K$}:
	\State $y^* \leftarrow \operatorname*{argmax}_{y \in V_{\setminus A}} \sigma_{y|A}^2 $ \quad using \eqn{eqn:gp:std}
	\State $A \leftarrow A \cup y^*$
\EndFor
\end{algorithmic}
\end{algorithm}
This will automatically choose sensors far away from each other, as illustrated in \fig{fig:gp}(b).
As a side effect, the selected locations tend to sit on the boundary of the space, making them in principle inefficiently using their full detection disk.

\subsubsection{Mutual Information}\label{sec:MI}

Another criterion suggested in~\cite{krause08near} is the Mutual Information (MI) which measures the shared information between selected and unselected locations:
\begin{align}
\mathrm{MI}(A, D_{\setminus A}) = H(D_{\setminus A}) - H(D_{\setminus A} | A) \label{eqn:MI}
\end{align}
Where we use here $D$ as the set of all locations (also those that are not permissive as sensor location (all light blue points in \fig{fig:filtering}(a)).
An intuitive explanation is given in \fig{fig:gp}(d).
Maximizing the MI between $D_{\setminus A}$ and $A$ is also a combinatorial problem,
such that a greedy method is used as well.
In each step we pick the location with maximum additive Mutual Information:
\begin{align}
y^* = \argmax_{y \in V_{\setminus A}} \left [ \mathrm{MI}(A \cup y, D_{\setminus {A \cup y}}) - \mathrm{MI}(A, D_{\setminus A})
\right]\label{eqn:MI:max}
\end{align}
as detailed in \alg{alg:mi}.

\begin{figure}
      \centering
      \includegraphics[scale=1]{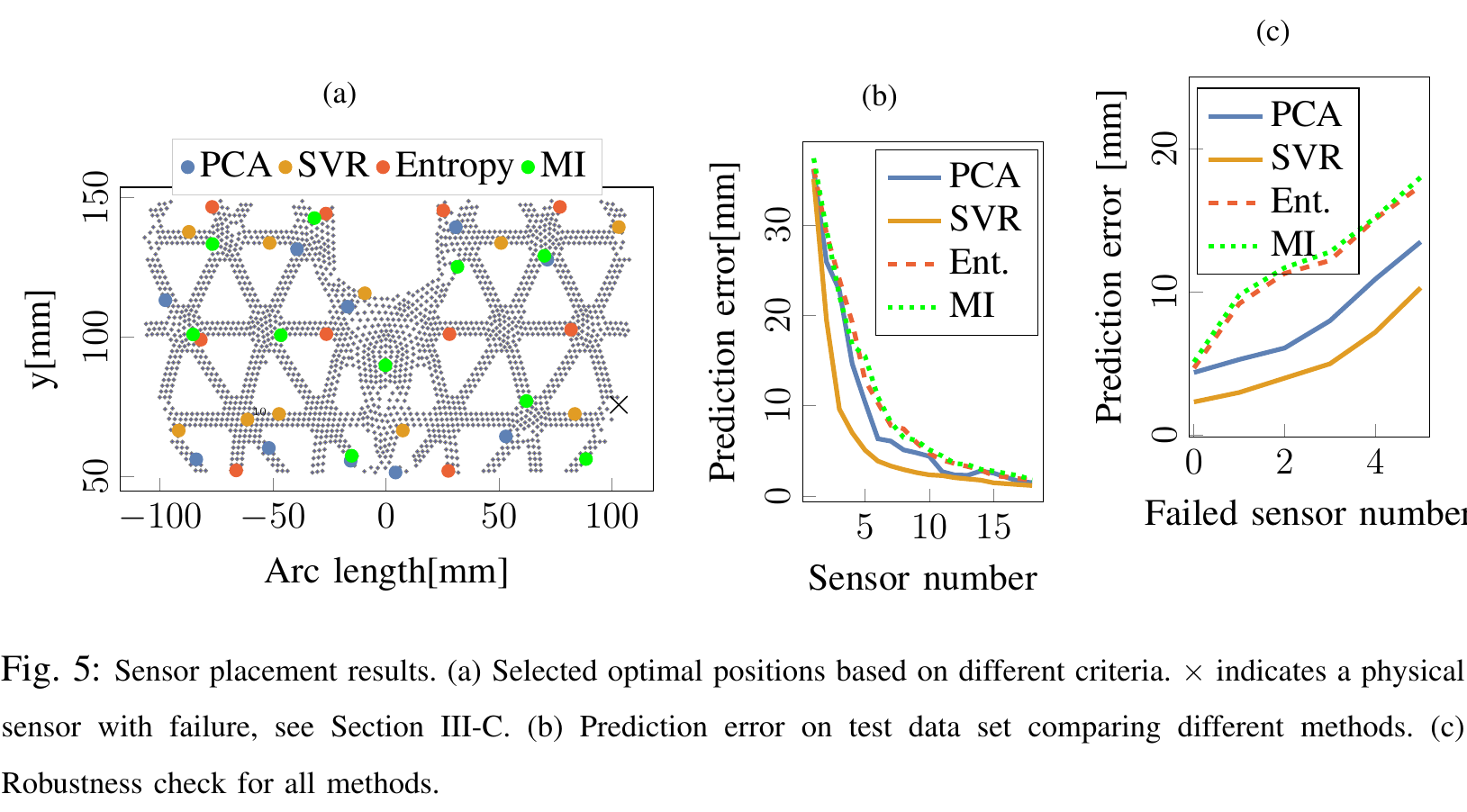}
      \caption{format = empty}
	  \label{fig:select}
\end{figure}

\begin{algorithm}
\caption{Mutual Information Criterion}\label{alg:mi}
\begin{algorithmic}[1]
\State {Input data: GP, sensor Budget $K$}{}
\For {$i:1$ \textbf{to} $K$}:
       \State $B(y) \leftarrow D \setminus(A \cup y)$
	\State $y^* \leftarrow \argmax_{y \in V_{\setminus A}} \left({\sigma_{y|A}^2} \over {\sigma_{y|B(y)}^2}\right) $\quad using \eqn{eqn:gp:std}
	\State $A \leftarrow A \cup y^*$
\EndFor
\end{algorithmic}
\end{algorithm}

\section{Results}\label{sec:results}
Using the methods presented above, we first determine the optimal sensor placement
based on the simulation results. Afterwards, we apply this to a real robotic limb.

\subsection{Optimal Sensor Placement and Validation in Simulation}\label{sec:placement}
Based on the preprocessed dataset acquired using finite element simulation (see \sec{sec:fes} and \ref{sec:filter}),
we compare the performance of the sensor placement using the different selection methods.

In \fig{fig:select}(a) we present the selected sensor positions.
We notice that the PCA-QR method places the sensors on the edges of each beam whose deformation has highest variances.
In contrast, the SVR method arranges sensors onto the two off-center parallel beams ($y \approx 70\,\mathrm{mm}$ and $y \approx 130\,\mathrm{mm}$), in which deformation from both sides can be measured.
The $10^{th}$ sensor position suggested by SVR is located closely to two already selected sensors rather than an expected more centered position.
However, as it turns out that the $11^{th} - 30^{th}$ sensor positions not shown in \fig{fig:select}(a) suggested by SVR are with high majority distributed closely to the upper and lower boundaries.
Because the areas near boundaries are more rigid and less sensitive to applied force.
More sensors are needed in these areas for high prediction precision. 
The model-based methods (Entropy, MI) are only based on the geometric information.
The Entropy criterion recommends the sensors to be placed toward the edges and homogeneously distributes them on the entire space.
The Mutual Information criterion suggests the positions to be more centered.

After picking the optimal sensor positions, we evaluate the four methods by comparing the prediction performance using the SVR force location inference (\sec{sec:inference}).
As shown in \fig{fig:select}(b), data-driven methods work generally better because they can exploit the structure of the data.
Note, that the PCA-QR method is not a greedy method and it suggests different combinations of sensors for each sensor budget $K$. For that reason, the prediction error is not guaranteed to be in descending order.

In general we notice that a small number of sensors can already lead to a relatively small prediction error (on unseen locations). Based on the simulation data we obtain a $<10\, \mathrm{mm}$ precision for 5 sensors,
selected using the greedy SVR method (\sec{sec:greedysvr}).
With 10 sensors the prediction performance is roughly $ 2.5\,\mathrm{mm}$

\subsubsection{Probing Robustness to Failure}\label{sec:robustness}
As any physical device is susceptible to failure, we also checked the robustness of the selected positions
 against failing sensors.
We tested the prediction error with different degrees of sensor failures, \ie out of 10 sensors $1$ to $5$ are broken.
\fig{fig:select}(c) presents the results.
The greedy SVR method has the highest robustness against sensor failures.
For $\nicefrac{3}{10}$ failures the performance drops by $5\,\mathrm{mm}$

\subsection{Experimental Evaluation of Hardware}\label{sec:hardware}

We use the 10 selected sensor positions by the greedy SVR method and implemented it on a hardware limb.
The next subsections introduce the sensor choice, the sensor assembly process, the data acquisition
 and finally the results of the force inference on the real system.

\subsubsection{Sensor choice}
In this section, we motivate the choice of the physical sensors.
We have chosen Strain Gauge (SG) sensors because they are generally cheap, widely available and relatively straightforward to use. 
Although we have also found cases of broken sensors after assembly.
In order to work with the strong deformations of the 3D-printed plastic robot parts, we select SGs with 20\% elongation rate. 
In our settings, the SG's finite extension is tolerant within the maximum elastic deformation of the limb.
These have a long lifetime and high fatigue strength.
One drawback of SG is that it only measures deformation along one direction.
For our beam-shaped structure, the SGs are assembled along the beam directions where the dominant deformation happens according to FEM.

Alternatively, we considered active sensors such as piezoelectric sensors~\cite{Liao05} transforming mechanical deformation into electrical energy and triboelectric effect based sensors~\cite{WANG2015436} charging through frictional contact. A typical problem of these sensors is that they cannot detect static loads.
In terms of passive sensors we considered capacitive ones, but they will react differently for conducting and consisting spatial limits.
Another interesting category is optical sensors like Fiber Bragg grating measuring a change in a deformed glass fiber based on the wavelength change of the reflected light.
The processing equipment for these sensors is rather involved, expensive and bulky.

\begin{figure}
      \centering
      \includegraphics[scale=1]{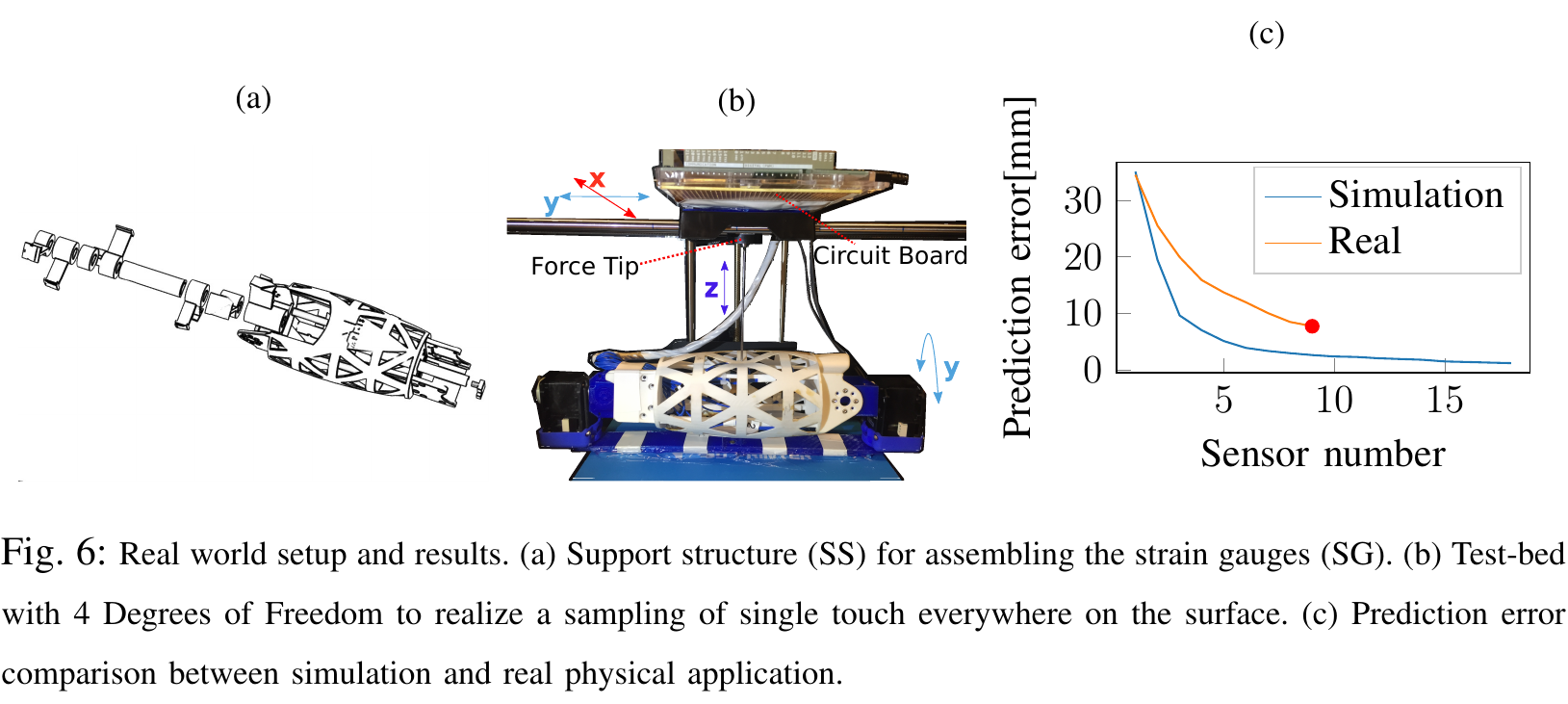}
      \caption{format = empty}
      \label{fig:hardware}
\end{figure}

\subsubsection{Sensor assembly}
The selected Strain Gauge (SG) sensor has to be attached to the inside of the 3D plastic structure
 with precaution in order to avoid damage or malfunction.
Since the SG only measure deformations along one direction, 
The assembly procedure is a bit challenging because it is inside the hollow object.
We developed an assembly method with a specific support structure (SS), as shown in \fig{fig:hardware}(a).
The SS has little arms to pre-tighten each of the sensors at the right position properly to the surface.
The arms are held in place by a middle axis during adhesive curing process then can be pulled out for disassembly.
The assembly process is described in Tab.~\ref{tab:assembly}
\begin{table}[h]
  \caption{Assembly procedure}\label{tab:assembly}
\label{table_example}
\begin{center}
\begin{tabular}{|c|c|}
\hline
Step & Details\\
\hline
1 & Wire SG and cover SG with scotch tape to isolate adhesive.\\
\hline
2 & Cover SS with preservative film to isolate adhesive.\\
\hline
3 & Insert absorbent wool between SG and SS to absorb adhesive.\\
\hline
4 & Position SG on SS.\\
\hline
5 & Clean internal surface of skeleton and SG surface.\\
\hline
6 & Coat SG and internal skeleton surface with prepared adhesive.\\
\hline
7 & Pre-tighten the whole structure and cure for $6h$ at $+20 ^\circ C$\\
\hline
8 & Disassemble SS and clean surface.\\
\hline
\end{tabular}
\end{center}
\end{table}

\subsubsection{Data Acquisition}

To acquire the dataset for training the machine learning algorithm, we have to record the
sensor measurements for many force applications.

\paragraph{Amplifier Circuit}

The conventional data acquisition circuit for SG is the Wheatstone bridge \cite{Stefanescu2011} which measures electrical resistance's change by balancing two legs of a bridge circuit.
As SG is sensitive not only to mechanical stress and but also to temperature variance, temperature compensation function has to be integrated into the circuit.

In our project, we adopt half-bridge Wheatstone, operational amplifier of MCP609 and Arduino Due.
The circuit has 12 I/O ports with 12 bits of resolution in which the SGs' deformation is amplified by a factor of 330 and converted to 4096 different values over $5\,\mathrm{V}$.

\paragraph{Test-bed}

To collect a large amount of force-measurements in an automated way, we designed test-bed
based on a modified 3D printer.
The printer offers 3 Degrees of Freedom (DoF) given by the Cartesian translation in $x, y, z$.
In addition, we add one axis of rotation.
In order to measure the forces the print-head of the printer is replaced by a force sensor tip (FC2231), see \fig{fig:hardware}(b).

\subsection{Experimental Results}\label{sec:results:hardware}

In this section, we validate the proposed \method{} on the modified limb of the Poppy robot~\cite{DasPoppy}.
The limb is designed to have an inside rigid support to ideally reduce the influence of forces at the joints and the support structure's elasticity, and a flexible shell to detect the touch.
We assembled 10 sensors according to the placement determined in \fig{fig:select}(a).
One of the sensors was malfunctioning, as indicated by a cross in the figure.
Each sensor value is calibrated to be zero if no force is applied.
While recording from the 9 remaining sensors, the force tip of the test-bed is stimulating the surface with different forces.
More specifically, the force tip is moved towards the structure to apply nominal force. As soon as a contact is registered, the
force tip is moved in small steps to a maximum penetration depth of $2\,\mathrm{mm}$.
In this way, different force magnitudes are obtained per location.
We collect data for 3000 locations on the surface, avoiding the edges and boundaries so that the force tip does not slide off.

From each location, we first use the largest force simulation and split the 3000 data points into 80\% training and 10\% for validation and test respectively.
After training the Support Vector Regression and performing hyper-parameter selection based on the validation set ($C= 20$, $\epsilon = 10^{-6}$ and $\gamma = 2 \times 10^{-3}$)
we evaluate the inference performance on the unseen force-locations.
\fig{fig:hardware}(c) presents the results compared to the simulations.
The hardware implementation achieves half of the simulation accuracy.
The average prediction precision of the force position is below $8\,\mathrm{mm}$ when using 9 sensors.
Given that our structure has a total surface of $200\,\mathrm{mm}\times120\,\mathrm{mm}$, this is a very high precision.

Besides, we train the SVR for different force amplitude, varying from light to strong touch, and report the prediction precision of the force information in different force intervals.
As shown in Tab.~\ref{tab:force prediction with intervals}, SVR has low prediction precision for light touch and high precision for strong touch.
Presumably because less sensors get activated by light touch.
The absolute prediction precision for the force's amplitude varies little \wrt force strength.
Consequently, any strong touches on the surface can be reliably detected and be used as a warning signal.
This improves the haptic system's robustness.

\begin{table}
	\caption{Prediction precision of force's postion and amplitude \wrt different force strength}\label{tab:force prediction with intervals}
	\begin{center}
		\begin{tabular}{|c|c|c|}
		\hline
		Force Interval & Position Error [mm] & Amplitude Error [N]\\
		\hline
		\hline
		0 - 4.9 $N$     & 25.43 $\pm$ 13.36  & 1.05 $\pm$ 1.01  \\
		4.9 - 9.8 $N$   & 11.95 $\pm$ 11.85  & 1.19 $\pm$ 1.19  \\
		9.8 - 19.6 $N$  & 5.90  $\pm$ 7.79   & 1.42 $\pm$ 1.78  \\
		19.6 - 34.3 $N$ & 4.48  $\pm$ 6.29   & 1.54 $\pm$ 2.21  \\
		\hline
		\end{tabular}
	\end{center}
\end{table}
\section{Discussion}\label{sec:discussion}
We present a method to obtain a robust haptic sensing system using only a small number of inexpensive deformation sensors.
The performance of the sensation device is powered by a machine learning approach.
After a learning period, the system can reliably localize touch all around a curved surface.
Apart from being inexpensive, the system is also very durable as the deformation sensors can be placed inside the structure.
Only a few ($\sim 10$) sensor values need to be acquired and processed.
The computational requirements are also low during operation as the
inference of the force location is done via Support Vector Regression. However, other machine learning methods, such as Deep Neural Networks, are feasible.

We also compared different methods of computing optimal sensor locations for a very sparse sensor configuration. We found that data-driven methods outperform geometry-based methods.
The right selection strategy can reduce the required number of sensors by 50\% without significant loss in precision.
In future work, we want to investigate multi-touch and accurate force magnitude prediction.

\section*{Acknowledgment}
The authors thank the International Max Planck Research School for Intelligent Systems (IMPRS-IS) and the China Scholarship Council (CSC) for supporting Huanbo Sun.

\bibliographystyle{IEEEtran}

\end{document}